\documentclass[twocolumn]{article}  
\usepackage{graphicx}
\usepackage{caption}
\usepackage{xcolor}
\usepackage{subcaption}
\usepackage{amsmath}
\usepackage{tikz}
\usetikzlibrary{positioning, arrows.meta}
\usepackage{physics}
\usepackage{pgfplots}
\usepackage[siunitx]{circuitikz}
\usepackage[outline]{contour}
\usetikzlibrary{decorations.markings}
\tikzset{>=latex}
\usepackage{adjustbox}
\usepackage{multicol}
\usepackage{amsfonts}
\usepackage{enumitem}
\usepackage{geometry}  
\title{Physics-Informed Neural Networks for Electrical Circuit Analysis: Applications in Dielectric Material Modeling}
\author{Reyhaneh Taj}
\date{}

\begin{document}

\twocolumn[
    \begin{@twocolumnfalse}
        \maketitle
        \begin{abstract}

    Scientific machine learning (SciML) represents a significant advancement in integrating machine learning (ML) with scientific methodologies. At the forefront of this development are Physics-Informed Neural Networks (PINNs), which offer a promising approach by incorporating physical laws directly into the learning process, thereby reducing the need for extensive datasets. However, when data is limited or the system becomes more complex, PINNs can face challenges, such as instability and difficulty in accurately fitting the training data. In this article, we explore the capabilities and limitations of the DeepXDE framework, a tool specifically designed for implementing PINNs, in addressing both forward and inverse problems related to dielectric properties. Using RC circuit models to represent dielectric materials in HVDC systems, we demonstrate the effectiveness of PINNs in analyzing and improving system performance. Additionally, we show that applying a logarithmic transformation to the current (\(\ln(I)\)) significantly enhances the stability and accuracy of PINN predictions, especially in challenging scenarios with sparse data or complex models. In inverse mode, however, we faced challenges in estimating key system parameters, such as resistance and capacitance, in more complex scenarios with longer time domains. This highlights the potential for future work in improving PINNs through transformations or other methods to enhance performance in inverse problems. This article provides pedagogical insights for those looking to use PINNs in both forward and inverse modes, particularly within the DeepXDE framework.


        \end{abstract}
        \vspace{0.2cm}
        
        
        \vspace{1cm}
    \end{@twocolumnfalse}
]

\section{Introduction}
The transition towards reducing greenhouse gas emissions globally necessitates significant changes in the electrical energy system. High Voltage Direct Current (HVDC) transmission is a powerful technology for efficiently transmitting electrical power across long distances. To ensure the reliability and optimization of HVDC insulation systems, it is essential to understand the dielectric properties of solid insulation materials. Most dangerous breakdowns are caused by the aging effects of HV insulation systems used within [HV] components, and there is still a lack of appropriate tools to diagnose such systems non-destructively and reliably in the field \cite{1}. Dielectric spectroscopy\cite{2} in the time domain plays a crucial role in providing key insights into these materials, particularly in applications such as HVDC converter transformers where liquid impregnated cellulose insulation is employed.

Machine learning (ML) has brought about a transformative paradigm shift in scientific practice. At the forefront of this revolution is scientific machine learning (SciML) \cite{3}, \cite{4}, \cite{5}. The primary objective of SciML is to intricately integrate established scientific knowledge with ML methodologies, creating robust ML algorithms that leverage our prior understanding to achieve greater efficacy. In recent years, the field has witnessed the emergence of a method known as Physics-Informed Neural Networks (PINNs) within the realm of artificial intelligence (AI) \cite{6}.

Diverging from conventional Neural Networks (NNs), which often demand datasets to yield accurate results, PINNs capitalize on an inherent grasp of physical principles, mitigating the need for extensive training data while bolstering computational efficiency. At the heart of PINNs lies Automatic Differentiation (AD), a pivotal mechanism enabling the network to compute derivatives concerning neural network parameters, input data, and biases. This automated derivation is pivotal for navigating the intricate landscape of partial differential equations (PDEs)\cite{7} and their associated boundary conditions, empowering the seamless integration of diverse physical models within the network architecture.

Representing a cutting-edge fusion of advanced physics principles, typically described by partial differential equations, with the adaptability of neural networks, PINN is a promising approach to addressing the complexities inherent in HV insulation systems.
\\
The objective of this project is to explore the capabilities and constraints of DeepXDE \cite{8}, a framework based on Physics-Informed Neural Networks (PINNs), in addressing both forward and inverse problems. In the forward mode, PINNs predict the behavior of current flow within the dielectric material, while in the inverse mode, they estimate the parameters of the physical equations that best describe the observed data. Additionally, we aim to analyze current measurements obtained during dielectric spectroscopy conducted in the time domain.
\\
The outline of this article is as follow; In Section 2, we provide a concise overview of Neural Networks (NN) followed by an in-depth description of Physics-Informed Neural Networks (PINNs), covering both the Forward Solution and Inverse discovery Mode. Section 3 delves into the description of our physical model formulations, wherein we illustrate and formulate the simplified physical system under study. Moving forward to Section 4, our focus shifts to the Implementation of DeepXDE, a crucial step in our methodology. In Section 5, we present the Results and Analysis derived from our investigation, providing insights into the performance and implications of our approach. Finally, in Section 6, we draw Conclusions and outline future directions, highlighting areas for further research and development in the field.

\section{Physics-Informed Neural Networks (PINNs)}

In the era of scientific machine learning (SciML), the fusion of established scientific knowledge with machine learning techniques is revolutionizing research methodologies. This transformation is prominently exemplified by the application of physics-informed neural networks (PINNs) to tackle partial differential equations (PDEs).
Differential equations (DEs) serve as fundamental tools for understanding how variables change in relation to each other over time. Enter physics-informed neural networks (PINNs): these innovative systems generate outputs that precisely match the behaviors prescribed by DEs, whether applied in physics, engineering, or other fields. Conversely, inverse physics-informed neural networks (iPINNs) operate in reverse, deducing the parameters of the underlying DEs based on observed responses. Both PINNs and iPINNs undergo training with a crucial constraint, ensuring that the neural network's input-output relationship precisely aligns with the DE being modeled, all achieved without the need for a large amount of data. This approach merges the power of neural networks with the precision of DEs, offering a versatile framework for modeling complex systems and phenomena across various disciplines.

In the following we will elucidate the concepts of PINNs with a focus on both the forward model and inverse model. 

Physics-Informed Neural Networks (PINNs) integrate deep learning with governing physical laws, making them ideal for problems where data is scarce. Unlike traditional neural networks that rely solely on large datasets, PINNs incorporate prior knowledge of differential equations, boundary, and initial conditions, reducing the need for extensive training data. This allows PINNs to efficiently model complex physical systems by encoding these constraints directly into the learning process.

\begin{figure}[ht]
    \includegraphics[width=1.0\columnwidth]{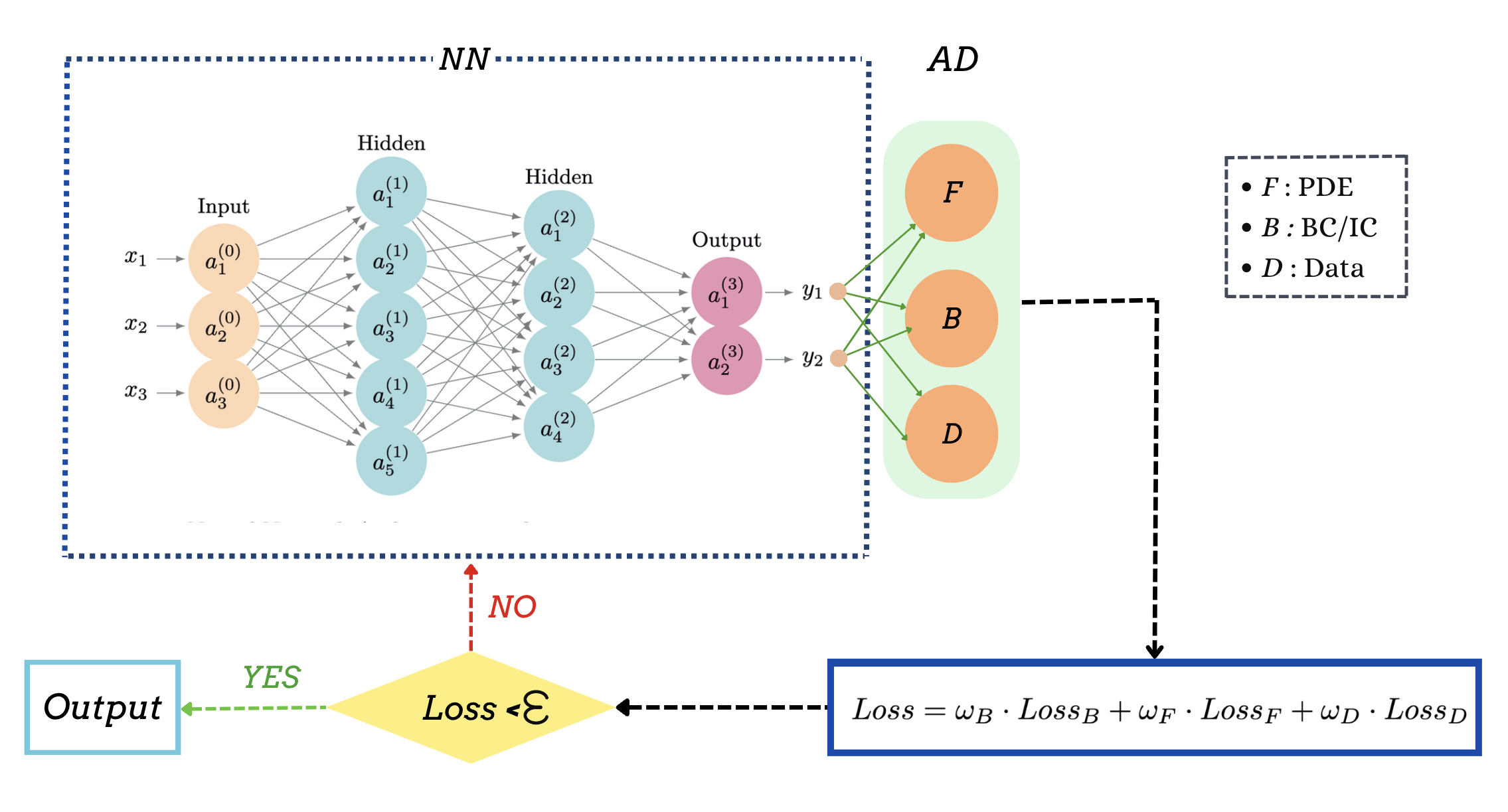}
    \caption{Overview of Physics-Informed Neural Networks (PINNs) architecture.}
    \label{fig:PINNs}
\end{figure}

In PINNs, the loss function includes not only data loss but also terms representing the residuals of differential equations and any boundary conditions. This enables the network to learn both from available data and from the underlying physics of the problem.

Figure \ref{fig:PINNs} illustrates the overall architecture of Physics-Informed Neural Networks (PINNs). In this structure, the input layer represents the variables of the system, such as spatial and temporal coordinates, while the output layer predicts the desired physical quantities, such as \(u(x,t)\), which may represent fields like temperature or current. The key innovation in PINNs is their ability to compute derivatives of these outputs with respect to the inputs using automatic differentiation, ensuring that the network respects the governing physical equations. The optimization process, as shown in the figure, involves minimizing a total loss function that includes both data loss and residuals from the physical model, ensuring the predicted outputs remain consistent with the system's behavior under physical constraints.

\subsubsection*{Forward Mode}

In forward problems, the goal is to approximate the solution \( u(z) \) for given parameters \( \lambda \). The loss function incorporates terms for both data and physics, ensuring the predicted solution respects the physical laws:

\[
L(\theta) = \omega_B L_B (\theta) + \omega_F L_F (\theta)
\]

where \( L_B \) accounts for boundary conditions and \( L_F \) ensures adherence to the governing differential equations.

\subsubsection*{Inverse Mode}

For inverse problems, where the goal is to estimate unknown physical parameters \( \lambda \), PINNs extend the loss function to include data discrepancies:

\[
L(\theta, \lambda) = \omega_B L_B (\theta, \lambda) + \omega_F L_F (\theta, \lambda) + \omega_D L_D (\theta, \lambda)
\]

This approach allows PINNs to handle noisy or limited data while ensuring the model's parameters are grounded in the physical reality of the system.

\section{RC Circuit Models for PINNs}

This section outlines the dynamic behavior of dielectric materials in RC circuits. We begin with a simple series RC circuit and gradually add complexity by introducing parallel RC configurations. Our objective is to derive the governing current differential equations for each configuration, along with their analytical solutions. 
These differential equations are embedded directly into the PINNs framework. During training, the network learns by minimizing a loss function that includes both data loss and residual terms from these physical equations. This ensures that the model not only fits the data but also adheres to the governing physical laws.

\subsection{Case 0: Single RC Circuit}

The simplest model is a series RC circuit. By applying Kirchhoff’s voltage law, we derive the following differential equation governing the current \(I(t)\):

\[
\frac{dI}{dt} + \frac{I}{RC} = 0, \quad I(t=0) = \frac{U_{\text{dc}}}{R}
\]

The analytical solution for the current is:

\[
I(t) = \frac{U_{\text{dc}}}{R} \exp\left(-\frac{t}{RC}\right)
\]

\subsection{Case 1: Two Parallel RC Circuits}

Next, we extend the model to include a second RC branch in parallel. The differential equation for the total current \(I(t)\) becomes:

\[
\frac{dI}{dt} + \frac{1}{R_1C_1}(I - \frac{U_{\text{dc}}}{R_0}) = 0
\]

The solution for the current in this case is:

\[
I(t) = \frac{U_{\text{dc}}}{R_0} + \frac{U_{\text{dc}}}{R_1} \exp\left(-\frac{t}{R_1C_1}\right)
\]

\subsection{Case 2: Three Parallel RC Circuits}

\begin{figure}[ht]
    \centering
    
    \begin{subfigure}{0.45\linewidth}  
        \centering
        \begin{circuitikz} [scale=0.8] 
            \draw (0,0) to[V, v=$U_{\text{dc}}$] (0,3)
            to[R, l=$R$, i= $I$] (3,3)
            to[C, l=$C$] (3,0)
            -- (0,0);
        \end{circuitikz}
        \caption{}
        \label{fig:case0}
    \end{subfigure}
    \hfill
    \begin{subfigure}{0.45\linewidth}  
        \centering
        \begin{circuitikz} [scale=1.0]
            \draw (0,4)  -- ++(1,0) coordinate (right) to[V, v=$U_{\text{dc}}$, i>=$I$] ++(3,0) coordinate (right);
            
            \draw (0,4) -- ++(0,-1.5) coordinate (left2) to[R, l=$R_0$, i=$I_{0}$] ++(3,0) coordinate (right2-end);
            
            \draw (0,4) -- ++(0,-3) coordinate (left3) to[R, l=$R_1$] ++(1.5,0) coordinate (right3) to[C, l=$C_1$, i=$I_{1}$] ++(2,0) coordinate (right3-end);
            
            \draw (0,4) -- ++(0,-4.5) coordinate (left4) to[R, l=$R_2$] ++(1.5,0) coordinate (right4) to[C, l=$C_2$, i=$I_{2}$] ++(2,0) coordinate (right4-end);
            
            \draw (right) |- (right2-end);
            \draw (right) |- (right3-end);
            \draw (right) |- (right4-end);
        \end{circuitikz}
        \caption{}
        \label{fig:case2}
    \end{subfigure}
    
    \caption{RC circuit configurations. (a) One R-C circuit in Case 0. (b) Three parallel RC branches in Case 2. This setup
can be generalized to Case \(N\) with additional RC branches
in parallel}
    \label{fig:rc_circuits}
\end{figure}
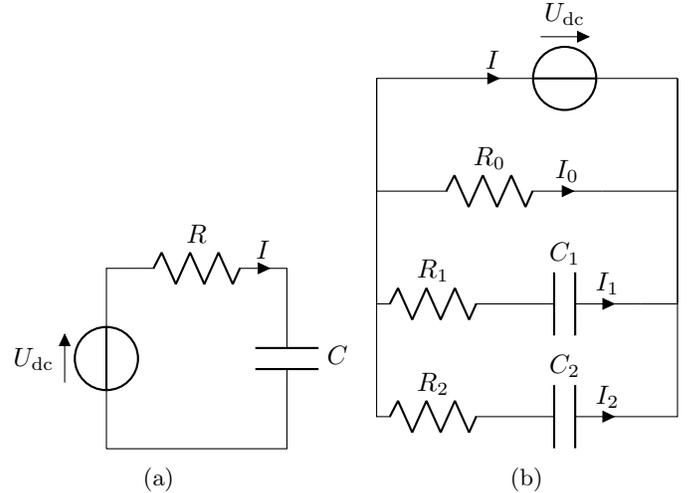

In this case, we add a third RC branch in parallel with the first two, as shown in Figure~\ref{fig:case2}. The current is given by:

\[
I(t) = I_{01}(t) + I_2(t)
\]

Where \(I_{01}(t)\) and \(I_2(t)\) are:

\begin{align*}
I_{01}(t) &= \frac{U_{\text{dc}}}{R_0} + \frac{U_{\text{dc}}}{R_1} \exp\left(-\frac{t}{R_1C_1}\right) \\
I_2(t)   &= \frac{U_{\text{dc}}}{R_2} \exp\left(-\frac{t}{R_2C_2}\right)
\end{align*}


\subsection{Generalization to Case \(N\)}

As seen in Figure~\ref{fig:case2}, the configuration for Case 2 involves three RC branches in parallel. This setup can be generalized to any number of parallel RC branches. The total current \(I(t)\) in the generalized case with \(N\) branches is given by:

\[
I(t) = \sum_{i=0}^{N} \frac{U_{\text{dc}}}{R_i} \exp\left(-\frac{t}{R_iC_i}\right)
\]

The generalization to \(N\) parallel RC branches is particularly important for real-world systems, where multiple components operate in parallel. This flexibility allows the model to handle more complex systems, making PINNs a versatile tool for analyzing a wide range of electrical circuits.

\subsection{Logarithmic Approach}

To improve numerical stability and reduce sensitivity, we can rewrite the current equations in terms of \(\ln(I)\). For example, the differential equation for Case 0 becomes:

\[
\frac{d\ln(I)}{dt} + \frac{1}{RC} = 0
\]

The logarithmic transformation of the current (\(\ln(I)\)) was chosen to improve numerical stability and performance. In complex systems or scenarios with limited data, large variations in current values can make the optimization process more difficult. By transforming the current into a logarithmic scale, the optimization becomes smoother, and the network is better able to converge on a stable solution.


As a summary, we have derived and summarized the differential equations and analytical solutions for different RC circuit configurations. These models will serve as the foundation for developing PINNs, enabling us to efficiently simulate and predict the behavior of complex RC systems.


\section{Implementation in DeepXDE}

DeepXDE, built on top of TensorFlow, facilitates the development of Physics-Informed Neural Networks (PINNs) by providing built-in functionalities to handle forward and inverse problems efficiently. The following sections describe how DeepXDE was implemented for predicting current (\(I\)) in electrical circuits.

\subsection{Forward Mode}

In forward mode, the goal is to predict the current (\(I\)) based on known system parameters. The steps include:

\begin{itemize}

    \item \textbf{Define the Computational Domain:} 
    DeepXDE discretizes the time domain (\(t\)) into specific points to serve as input variables, influencing prediction accuracy and granularity.
    
   \item \textbf{Specify the PDE:} 
   The governing differential equation (PDE) is defined using TensorFlow syntax, incorporating variables like \(R_i\) and \(C_i\) as system parameters. The PDE contributes to the total loss function during training.

    \item \textbf{Set Initial Conditions:} 
    For this study, we used initial conditions (IC) to define the system's starting state. DeepXDE supports common boundary conditions, though only ICs were used in this case.
    
    \item \textbf{Create Training and Testing Points:} 
    DeepXDE combines the geometry, PDE, and ICs to generate training points (collocation points) within the time domain. These points are used to train the network, with analytical solutions serving as references.
    
    \item \textbf{Construct the Neural Network:} 
    The network architecture (layers, activation functions, neurons) is specified using DeepXDE’s streamlined tools, ensuring it fits the problem’s complexity.
    
    \item \textbf{Set Hyperparameters and Train the Model:} 
    Hyperparameters such as learning rate and optimizer are set. The model is then trained to minimize the loss function:
    \[ Loss_{PINNs} = Loss_{PDE} + Loss_{IC} \]
    
    \item \textbf{Predict the PDE Solution:} 
    After training, the model predicts the desired PDE solution, which is evaluated against test data or analytical solutions to assess performance.
    
\end{itemize}

\subsection{Inverse Mode}

In inverse mode, the goal is to estimate unknown system parameters (e.g., resistances \(R_i\) and capacitances \(C_i\)) based on observed current (\(I\)) data over time. The steps include:

\begin{itemize}

    \item \textbf{Specify the Computational Domain:} 
    Similar to forward mode, the time domain over which current is observed is discretized, significantly influencing the model's ability to predict parameters accurately.

    \item \textbf{Define Variables and the Inverse Problem:} 
    The system’s unknown parameters (\(R_i\), \(C_i\)) are specified, and the inverse problem scope is defined based on the system’s differential equations.

    \item \textbf{Use Synthetic or Measured Data:} 
    Synthetic data is generated using assumed parameter values, while measured data comes from real-world observations. Both types of data are used for training and testing the model.

    \item \textbf{Set Initial Conditions and Generate Points:} 
    Training points are generated based on the geometry, PDE, initial conditions, and data. These points guide the model in learning the relationship between input data and physical parameters.
    
    \item \textbf{Construct the Neural Network:} 
    The architecture of the neural network is specified as in forward mode, but tailored to the inverse problem’s requirements.

    \item \textbf{Set Hyperparameters and Train the Model:} 
    The model is trained to minimize the following loss function, which includes terms for the PDE, data, and initial conditions:
    \[ Loss_{PINNs} = Loss_{PDE} + Loss_{Data} + (Loss_{IC}) \]

    \item \textbf{Predict System Parameters:} 
    After training, the model predicts the unknown system parameters (\(R_i\), \(C_i\)) based on the observed current data. The predictions are validated against analytical solutions to assess accuracy.
    
\end{itemize}

\section{Results}

This section presents the results of applying DeepXDE to solve forward and inverse problems in the context of modeling dielectric materials. We focus on evaluating the accuracy and stability of PINNs when predicting the current \(I\) in various circuit configurations, as well as their ability to estimate system parameters, such as resistance and capacitance, in the inverse mode. Particular attention is given to the role of the logarithmic transformation, which significantly improves the performance of PINNs in challenging scenarios involving complex circuits and limited data.

\subsection{Forward Mode: Current Prediction}

In the forward mode, PINNs are used to predict the current \(I\) across various circuit configurations. Figure~\ref{fig:Case0,I} shows the predicted current for Case 0, a simple RC circuit, demonstrating excellent agreement between the model’s predictions and the analytical solution over the 10-second time domain. The high accuracy and stability observed in this case serve as a baseline for evaluating the model's performance in more complex scenarios.

\begin{figure}[ht]
    \centering
    \includegraphics[width=0.49\textwidth]{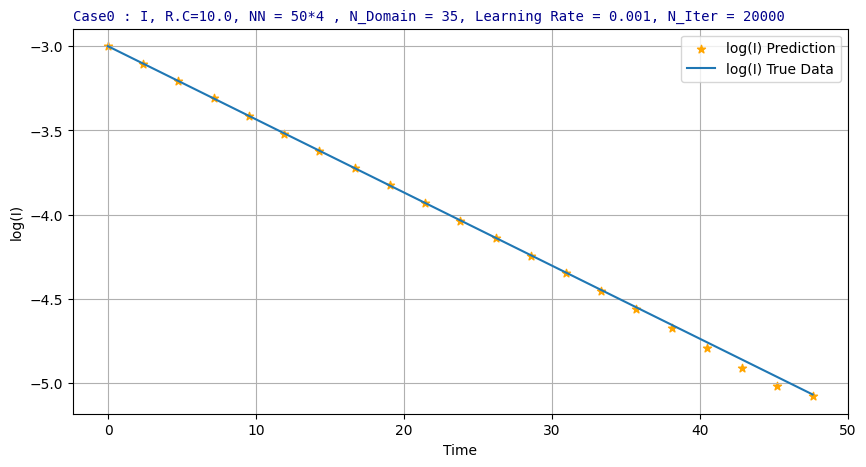}
    \caption{Predicted current for Case 0 as a function of time in forward mode. 'True data' refers to the analytical solution.}
    \label{fig:Case0,I}
\end{figure}

\begin{figure}[ht]
    \centering
    \begin{subfigure}{0.48\textwidth}
        \centering
        \includegraphics[width=\textwidth]{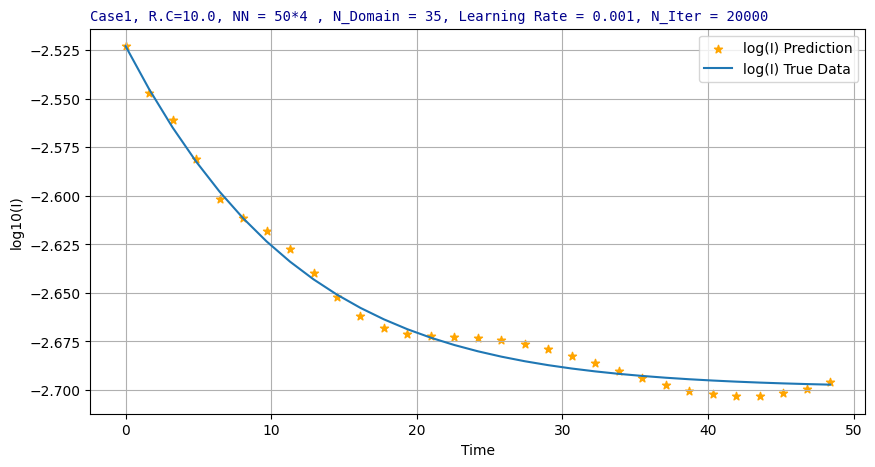}
    \end{subfigure}
    \caption{Predicted $\ln(I)$ with respect to time for Case 1. 'True data' refers to the analytical solution.}
    \label{fig:case1}
\end{figure}

However, as circuit complexity increases, discrepancies begin to emerge between the model's predictions and the analytical solution. Figure~\ref{fig:case1} presents the predicted current for Case 1, a more complex circuit involving two parallel RC branches. Although the model captures the overall behavior of the system, deviations from the true data are noticeable, especially beyond 2 seconds. These discrepancies highlight the model's increasing difficulty in handling more complex configurations, motivating the need for additional techniques, such as the logarithmic transformation, to improve stability and accuracy.

In Figure \ref{fig:case1}, we observe deviations in the current prediction beyond 2 seconds, revealing the model's difficulty in maintaining accuracy. This motivated the need for a more robust approach, such as the logarithmic transformation, which proved effective in stabilizing predictions for more complex cases.

\subsection{Logarithmic Transformation for Enhanced Predictive Performance}

To address the challenges observed in more complex circuits, we applied a logarithmic transformation to the current (\(\ln(I)\)). This transformation was introduced to improve the model’s stability and accuracy, particularly in scenarios where data is sparse or the system exhibits large variations in current over time. By transforming the current into a logarithmic scale, the optimization process becomes smoother, and the model is better able to handle the wide range of values that can arise in complex circuits.

\begin{figure}[ht]
    \centering
    \begin{subfigure}{0.48\textwidth}
        \centering
        \includegraphics[width=\textwidth]{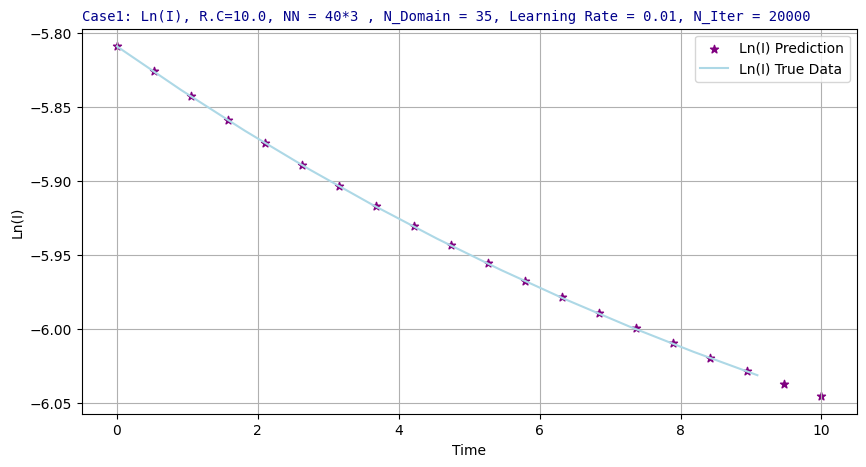}
        \subcaption{Case 1}
    \end{subfigure}
    \begin{subfigure}{0.48\textwidth}
        \centering
        \includegraphics[width=\textwidth]{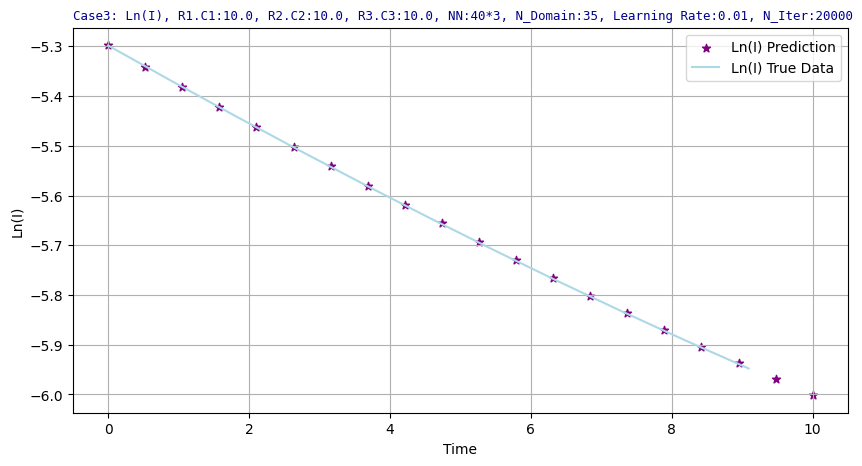}
        \subcaption{Case 3}
    \end{subfigure}
    \caption{Predicted $\ln(I)$ with respect to time for Case 1 (a), Case 3 (b). 'True data' refers to the analytical solution. The logarithmic transformation significantly improves prediction accuracy for more complex cases.}
    \label{fig:lnI_predictions}
\end{figure}

Figure~\ref{fig:lnI_predictions} (a) , (b) present the predicted \(\ln(I)\) for Case 1 and Case 3, respectively. In Case 1 (Figure~\ref{fig:lnI_predictions} (a)), we observe that the logarithmic transformation significantly reduces the discrepancies seen in the raw current predictions (as shown earlier in Figure~\ref{fig:case1}). The model now converges more closely to the true data, particularly beyond 2 seconds, where the original model struggled. Similarly, in Case 3 (Figure~\ref{fig:lnI_predictions} (b)), the transformation allows the model to maintain stability across the entire time domain, providing predictions that are consistent with the analytical solution.

These results demonstrate the power of the logarithmic transformation in stabilizing the optimization process and improving accuracy in complex circuit configurations. By mitigating the large variations in current values, the transformation enhances the PINN’s ability to converge on a stable solution, even when the data is sparse or the system becomes more intricate.




This transformation was a key result of this investigation, demonstrating that the logarithmic approach plays a crucial role in improving prediction accuracy in both forward and inverse modes. The ability of the model to converge more accurately when dealing with complex circuits highlights the power of this method.

\subsection{Hyperparameter Optimization and Time Domain Challenges}

A key aspect of improving the stability and accuracy of PINNs in this study was optimizing the model’s hyperparameters. We performed extensive experimentation to find the optimal combination of learning rate, collocation points, and network architecture, ensuring the model was able to predict current accurately in both simple and complex circuits. 

The learning rate was set to 0.01, balancing the need for efficient convergence with the risk of overshooting during optimization. The number of collocation points, crucial for capturing the system’s behavior, was set to 35 per 10 seconds in the time domain. This choice provided a sufficient resolution to accurately model the current, while keeping computational costs manageable. Additionally, the architecture of the neural network was optimized with 3 layers and 40 neurons per layer, chosen based on iterative experiments aimed at preventing overfitting and improving generalization.

Despite these optimizations, challenges arose when extending the time domain in more complex cases. For example, in Case 2, where the time domain was extended from 10 to 100 and 300 seconds, the model required a significant increase in the number of collocation points to maintain accuracy. In the 300-second case, the number of collocation points had to be increased tenfold to 1,050 points. Even with these adjustments, predictions became less reliable at larger time scales, highlighting the limitations of the model in maintaining accuracy over extended time domains.

These results show that while hyperparameter tuning improved model performance in shorter time domains, the model struggled to maintain consistency as the time domain expanded. One potential solution is to use adaptive collocation point sampling, where more points are dynamically allocated to areas with higher error. This could help manage the computational complexity while improving accuracy in longer time domains.

\subsection{Inverse Mode: Parameter Estimation Results}

In the inverse mode, the objective is to estimate system parameters, such as resistance (\(R\)) and capacitance (\(C\)), from observed current profiles. Using PINNs, we trained the model to learn the relationship between the synthetic current data and the underlying physical parameters governing the system.

\begin{figure}[ht]
    \centering
     \begin{subfigure}{0.48\textwidth}
        \centering
        \includegraphics[width=\linewidth]{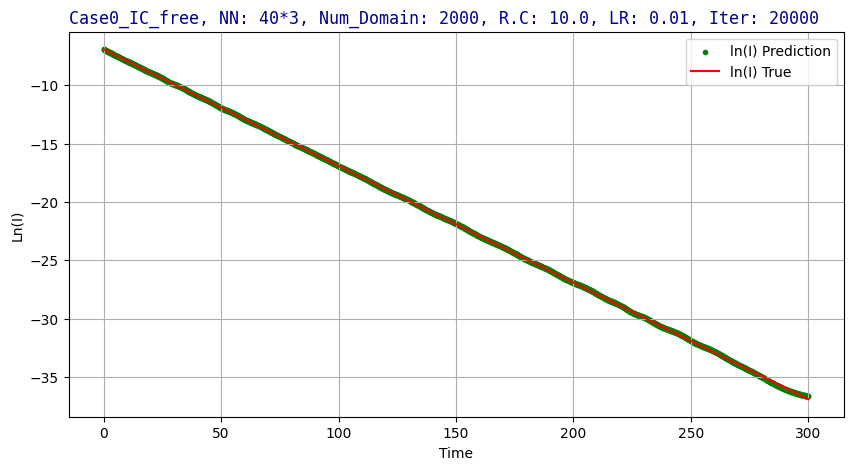}
        \caption{Case 0}
        \label{fig:case0a}
    \end{subfigure}\hfill
    \begin{subfigure}{0.48\textwidth}
        \centering
        \includegraphics[width=\linewidth]{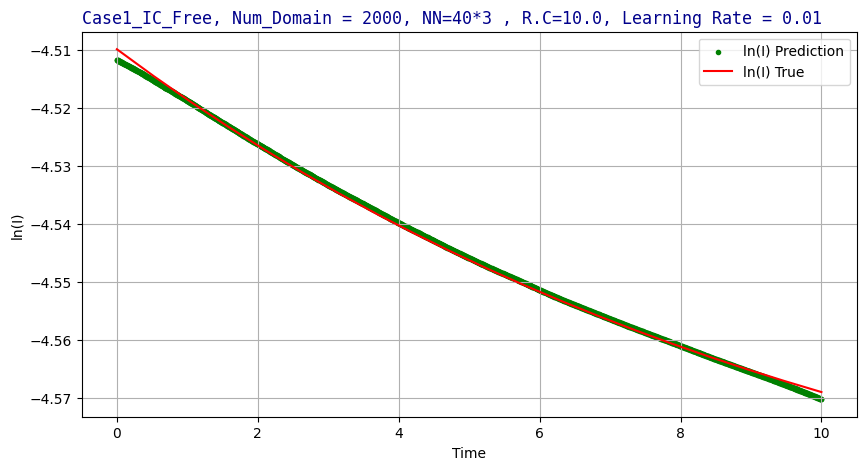}
        \caption{Case 1}
        \label{fig:case0b}
    \end{subfigure}
    \caption{Predicted current for Case 0, 1 in inverse mode using estimated parameters. The model shows high accuracy in parameter estimation for simple circuits.}
    \label{fig:case0_icf}
\end{figure}

For Case 0, the model performed well, accurately predicting both resistance and capacitance with minimal error. Figure~\ref{fig:case0_icf} shows the predicted current for Case 0 using the estimated parameters. For simple circuits, the predictions closely match the analytical solution, demonstrating the effectiveness of the inverse mode in estimating system parameters.

However, as the complexity of the circuit increased, the model faced greater challenges in estimating parameters accurately. In Case 1, where the circuit includes additional parallel branches, the error in the estimated parameters increased, particularly when the time domain was extended. These challenges are evident in Figure~\ref{fig:case0_icf}, where deviations from the true data become more noticeable as the time domain extends beyond 10 seconds. Despite these difficulties, the logarithmic transformation of the current helped improve the model’s stability, resulting in more accurate predictions compared to the raw current predictions.

One limitation we encountered was the sensitivity of the model to initial conditions and the need for careful tuning of the hyperparameters. In particular, as the time domain increased, the model struggled to maintain accuracy without increasing the number of collocation points significantly. This highlights the need for further refinement of the inverse mode, especially for more complex circuits and longer time domains.



Despite these challenges, the inverse mode of PINNs shows great potential for parameter estimation in complex systems. By further refining the model and exploring techniques such as adaptive sampling, it may be possible to achieve more accurate parameter estimation across a wider range of circuit configurations and time scales.




\section{Summary and Conclusion}

This study explored the application of Physics-Informed Neural Networks (PINNs) to both forward and inverse problems related to dielectric materials in HVDC systems. We demonstrated the effectiveness of PINNs in predicting the behavior of current in various circuit configurations and estimating system parameters, such as resistance and capacitance, from observed data. A key contribution of this work was the introduction of a logarithmic transformation to the current (\(\ln(I)\)), which significantly improved the model’s stability and accuracy, particularly in more complex circuits and data-sparse scenarios.

The results from the forward mode showed that PINNs can accurately predict current for simple circuits like Case 0. However, as the complexity of the circuits increased, such as in Case 1 and Case 3, discrepancies between the model’s predictions and the analytical solutions were observed, especially over extended time domains. The logarithmic transformation helped stabilize the optimization process, allowing the model to converge more closely to the true data and mitigating the discrepancies seen in raw current predictions.

In the inverse mode, we used PINNs to estimate key system parameters, such as resistance and capacitance. While the model performed well in simpler circuits, the accuracy of parameter estimation declined as the complexity increased. Sensitivity to initial conditions and hyperparameters further complicated the process, especially for longer time domains. Although the logarithmic transformation improved the stability of the model in some cases, significant challenges remained in obtaining accurate parameter estimations. This suggests that further work, potentially through additional transformations or adaptive sampling techniques, is needed to enhance the performance of PINNs in inverse problems.

This study not only showcases the potential of PINNs for solving forward problems, but also highlights the challenges and opportunities in using them for inverse problems. Although we did not achieve strong results in the inverse case, this area offers promising potential for future work. Exploring alternative transformations and sampling strategies could further improve the accuracy and robustness of the model.

In addition to its technical contributions, this work serves as a valuable pedagogical example for those interested in using PINNs and the DeepXDE framework. The results demonstrate the practical utility of these tools in dielectric research, providing a foundation for future studies to build upon.

The application of these methods shows great potential for improving the design, analysis, and performance of dielectric materials in HVDC systems. By accurately modeling the behavior of electrical circuits and estimating system parameters, PINNs offer a powerful tool for optimizing the performance of complex systems.

Future work should focus on refining the inverse mode of PINNs, particularly by exploring adaptive sampling techniques to enhance the model’s accuracy in complex circuits and over longer time domains. Further research into handling sparse data and optimizing hyperparameters could also improve the robustness of the model in real-world applications. By addressing these challenges, PINNs could become even more effective in analyzing and improving the performance of electrical systems.

\section*{Acknowledgements}

This paper is a result of my internship at Hitachi Energy Research Center, under the guidance and supervision of Olof Hjortstam and Tor Laneryd. I would like to thank them for their invaluable support, insights, and mentorship throughout this research project. Their expertise and encouragement greatly contributed to the development of this work.



\begin{thebibliography}{9}
\bibitem{1}
W.S. Zaenglaengl, "Dielectric spectroscopy in time and frequency domain for HV power equipment. I. Theoretical considerations", \emph{IEEE Electrical Insulation Magazine}, Volume 19, Issue 5, Sept.-Oct. 2003.

\bibitem{2}
F. Kremer and A. Schönhals, \emph{Broadband dielectric spectroscopy}, 2002.



\bibitem{3}
Nathan Baker, Frank Alexander, Timo Bremer, Aric Hagberg, Yannis Kevrekidis, Habib Najm, Manish Parashar, Abani Patra, James Sethian, Stefan Wild, Karen Willcox, and Steven Lee. Workshop Report on Basic Research Needs for Scientific Machine Learning: Core Technologies for Artificial Intelligence. Technical report, USDOE Office of Science
(SC) (United States), feb 2019. 

\bibitem{4}
Tony Hey, Keith Butler, Sam Jackson, and Jeyarajan Thiyagalingam. Machine learning and big scientific data. Philosophical Transactions of the Royal Society A: Mathematical, Physical and Engineering Sciences, 378(2166), mar 2020. ISSN 1364503X. doi: 10.1098/rsta.2019.0054.

\bibitem{5}
Salvatore Cuomo, Vincenzo Schiano di Cola, Fabio Giampaolo, Gianluigi Rozza, Maizar Raissi, and Francesco Piccialli. Scientific Machine Learning through Physics-Informed Neural Networks: Where we are and What’s next. arXiv, jan 2022.

\bibitem{6}
M. Raissi et al., "Physics-informed neural networks: A deep learning framework for solving forward and inverse problems involving nonlinear partial differential equations", \emph{Journal of Computational Physics}, Volume 378, 2019, Pages 686-707.
\bibitem{7}
Blechschmidt,J.,Ernst,O.G.:Three ways to solve partial differential equations with neural networks– A review. GAMM-Mitteilungen 44(2), e202100,006 (2021).

\bibitem{8}
Lu Lu, Xuhui Meng, Zhiping Mao, and George Em Karniadakis, "DeepXDE: A Deep Learning Library for Solving Differential Equations", SIAM ReviewVol. 63, Iss. 1 (2021)10.1137/19M1274067.


\bibitem{9}
Goodfellow I, Bengio Y, Courville A. Deep Learning. MIT Press; 2016.

\end{thebibliography}
\end{document}